\documentclass{article}

\usepackage[preprint]{neurips_2025}

\usepackage[T1]{fontenc}
\usepackage[table,dvipsnames]{xcolor}  
\usepackage{hyperref}
\usepackage{url}
\usepackage{booktabs}
\usepackage{multirow}
\usepackage{array}
\usepackage{adjustbox}
\usepackage{amsmath}
\usepackage{amsfonts}
\usepackage{nicefrac}
\usepackage{microtype}
\usepackage{colortbl}
\usepackage{graphicx}
\usepackage[most]{tcolorbox}
\usepackage{subfigure}
\usepackage{wrapfig}

\title{Long or short CoT? Investigating Instance-level Switch of Large Reasoning Models}

\author{%
  Ruiqi Zhang, Changyi Xiao,  Yixin Cao\thanks{Corresponding author} \\
  School of Computer Science, Fudan University \\
  \texttt{24210240070@m.fudan.edu.cn}, \texttt{\{changyi\_xiao, yxcao\}@fudan.edu.cn}
}
\date{} 

\begin{document}

\maketitle

\begin{abstract}


With the rapid advancement of large reasoning models, long Chain-of-Thought (CoT) prompting has demonstrated strong performance on complex tasks. However, this often comes with a significant increase in token usage. In this paper, we conduct a comprehensive empirical analysis comparing long and short CoT strategies. Our findings reveal that while long CoT can lead to performance improvements, its benefits are often marginal relative to its significantly higher token consumption. Specifically, long CoT tends to outperform when ample generation budgets are available, whereas short CoT is more effective under tighter budget constraints. These insights underscore the need for a dynamic approach that selects the proper CoT strategy based on task context and resource availability. To address this, we propose \textbf{SwitchCoT}, an automatic framework that adaptively chooses between long and short CoT strategies to balance reasoning accuracy and computational efficiency. Moreover, SwitchCoT is designed to be budget-aware, making it broadly applicable across scenarios with varying resource constraints. Experimental results demonstrate that SwitchCoT can reduce inference costs by up to 50\% while maintaining high accuracy. Notably, under limited token budgets, it achieves performance comparable to—or even exceeding—that of using either long or short CoT alone.

\end{abstract}

\section{Introduction}






Recently, large reasoning models (LRMs), such as DeepSeek-R1 \citep{DeepSeekAI2025DeepSeekR1IR}, OpenAI-O1 \citep{OpenAI} have achieved remarkable success by generating explicitly long chains of thought (CoT), often significantly exceeding the length of depth of reasoning produced by traditional large language models. Recent studies have explored long CoT from two key perspectives. First, although long CoT can enhance reasoning quality, it substantially increases computational cost and may introduce issues such as overthinking \citep{Chen2024DoNT}. These findings raise important questions about whether long CoT is necessary for all tasks. Second, several works have found that long CoT is not always superior to short CoT \citep{Wu2025WhenMI, Ma2025ReasoningMC}, showing that in certain scenarios, short CoT can be more efficient and robust.

To gain a deeper understanding, we conduct a systematic, instance-level analysis comparing long and short CoT strategies from multiple perspectives. Our findings show that while long CoT can offer substantial performance improvements in certain cases, it often yields only marginal gains relative to its significantly higher token consumption. These insights underscore the importance of a dynamic strategy that adapts the choice between long and short CoT based on task requirements and computational constraints.



Building on our empirical findings, we propose a new method called \textbf{SwitchCoT}, which enables large reasoning models to automatically switch between generating long and short CoT at the instance level. SwitchCoT strikes a balance between reasoning effectiveness and computational efficiency. Furthermore, it is explicitly designed to be budget-aware, making it well-suited for scenarios where computational resources are constrained. 

In practice, resource constraints are common, and effective reasoning under limited budgets is crucial. Additionally, the optimal reasoning strategy for a given instance may vary as the available budget changes. SwitchCoT addresses this challenge by dynamically adapting its reasoning approach based on both the instance and the current budget, ensuring more flexible and cost-effective performance.

Our approach is inspired by the dual-process theory in cognitive science, which posits that humans alternate between fast, intuitive thinking and slow, analytical reasoning depending on the context. Similarly, some instances are best handled through concise reasoning (short CoT), while others require more detailed analysis (long CoT). SwitchCoT emulates this adaptive capability, allowing LRMs to select the most suitable reasoning mode for each scenario.

We finally conduct experiments to demonstrate the substantial advantages of SwitchCoT. Results show that it can significantly reduce token consumption—by up to 50\%—on mathematical and sentiment analysis tasks. Importantly, this efficiency gain does not come at the expense of performance: SwitchCoT matches or even surpasses the accuracy of long CoT on various tasks, including outperforming it on knowledge-intensive and sentiment tasks. Our codes are available at \href{https://github.com/troublemaker5213/SwitchCoT}{https://github.com/troublemaker5213/SwitchCoT}

In summary, our key contributions are:

\begin{enumerate}
    \item We present a comprehensive empirical analysis showing that while long CoT generally improves accuracy, short CoT is more efficient and preferable in resource-constrained settings—highlighting the need for adaptive strategy selection.
    \item We introduce \textbf{SwitchCoT}, a budget-aware framework that enables LRMs to dynamically switch between long and short CoT strategies, balancing effectiveness and efficiency.
    \item Experimental results demonstrate that \textbf{SwitchCoT} reduces token usage by up to 50\% while maintaining or even improving accuracy, showcasing its practical utility.
\end{enumerate}



\section{Related work}

\subsection{Chain-of-Thought}

CoT reasoning, introduced by \citep{NEURIPS2022_9d560961}, improves the performance of large language models by explicitly generating explicit intermediate reasoning steps prior to producing final answers. Building on this idea, subsequent research has sought to optimize CoT for more complex tasks. Tree-of-Thought \citep{yao2023tree} extends CoT by treating coherent text segments as units of exploration, allowing the model to navigate multiple reasoning paths through techniques like backtracking and lookahead. Graph-of-Thought \citep{besta2024graph} further generalizes this approach by structuring model-generated information as a graph, effectively capturing dependencies and enabling collaborative reasoning. OpenAI-O1 \citep{OpenAI} and DeepSeek-R1 \citep{DeepSeekAI2025DeepSeekR1IR} introduce an explicit deliberation phase---referred to as long CoT---prior to generating reasoning steps, further enhancing performance on challenging tasks by mimicking human-like reflection. In this work, we conduct a comparative study of long and short CoT strategies, aiming to understand their differences and develop a method that adaptively selects the optimal reasoning strategy for each instance.



%


\subsection{Efficient Reasoning}


LRMs have demonstrated improved performance on complex tasks by introducing an explicit "thinking" phase prior to answer generation---an advancement over traditional CoT reasoning. However, this more elaborate reasoning process often incurs significant computational overhead, reducing overall efficiency. As a result, recent research has increasingly focused on enhancing reasoning efficiency, aiming to retain the performance benefits of long CoT while mitigating its resource costs. For instance, \citep{ma2025reasoning} eliminates the reasoning phase entirely by directly generating multiple candidate answers and aggregating them. \citep{liu2025thought} injects thought sequences from smaller models into designated “thinking” tags to guide reasoning. \citep{xia2025tokenskip} reduces output length by identifying and removing low-importance tokens, while \citep{liu2024can} iteratively refines intermediate steps to yield more concise CoT outputs. In parallel, several studies propose prompt-based strategies \citep{han2024token, lee2025well, renze2024benefits, xu2025chain}, and others explore compressing latent representations to lower inference costs \citep{cheng2024compressed, hao2024training, su2025token}. Despite these advances, most existing methods focus on optimizing or compressing generated reasoning content. In contrast, the strategy of input-side selection—dynamically deciding whether to use short or long CoT based on the characteristics of each input—remains relatively underexplored.


\section{Empirical Analysis on Long and Short CoT }
\label{sec:empirical}


To guide the design of subsequent methods and experiments, we first summarize several key empirical findings, followed by extensive validation:

\begin{itemize}
\item Long CoT yields significant improvements on tasks that require explicit logical reasoning, but may introduce redundancy or noise in tasks where such reasoning is unnecessary.
\item The performance benefit of long CoT increases with task difficulty, underscoring its effectiveness in addressing complex reasoning challenges.
\item As model size increases, the relative advantage of long CoT diminishes, suggesting that larger models are often capable of achieving strong performance with short CoT strategies.
\item The cost-effectiveness of additional reasoning varies across domains---more tokens do not uniformly result in higher accuracy.
\end{itemize}

These observations indicate that relying on a single, fixed reasoning strategy is suboptimal, especially across tasks with diverse reasoning requirements and computational constraints.

\subsection{Experimental Settings}

This subsection outlines our experimental setting and the reasoning control method. 

\paragraph{Models \& Method}

LRMs typically default to generating long CoT, which poses challenges for fairly evaluating the effectiveness of short CoT strategies. To enable controlled comparisons, we surveyed widely adopted LLMs that support configurable reasoning styles and selected those capable of explicitly generating both short and long CoT through prompt manipulation. Our study includes models from the DeepSeek distillation series (\citealp{DeepSeekAI2025DeepSeekR1IR})—specifically, DeepSeek-R1-Distill-Qwen-7B, 14B, and 32B, as well as DeepSeek-R1-Distill-LLaMA-8B—and Qwen3-8B (\citealp{qwen3}).


By default, the model generates an explicit reasoning stage enclosed within \textit{<think>} and \textit{</think>} tags. To elicit short CoT behavior, we insert a placeholder \textit{<think>} block containing content-free text (e.g., "Okay, I think I have finished thinking."), effectively signaling the model to minimize reasoning without omitting the required structure.

\paragraph{Datasets}




We conducted experiments on a diverse set of datasets spanning various task types, including mathematical reasoning, knowledge-based reasoning, social reasoning, factual question answering, sentiment analysis, and creative matching tasks. The full list of the datasets is provided in Appendix~\ref{app:dataset}.

\begin{figure}
    \centering
    \includegraphics[width=1\linewidth]{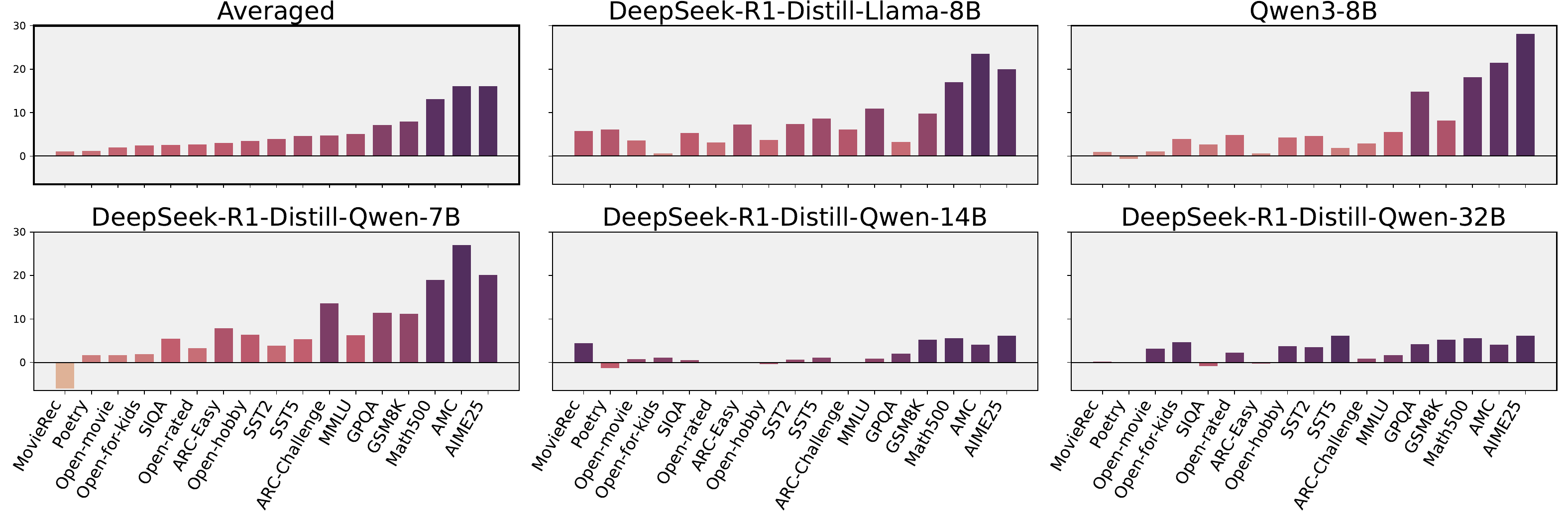}
    \caption{\textbf{Performance gains} from using long CoT across different datasets and representative models. Datasets are ordered by average accuracy. The highest accuracies are observed in mathematical reasoning tasks (AIME25, AMC23, MATH500, GSM8K), followed by knowledge-based tasks (GPQA, MMLU, ARC), both of which generally involve higher reasoning complexity.}
    \label{fig:improve}
\end{figure}


\subsection{Analysis on Three Dimensions}

Motivated by human intuition—specifically, the tendency to invest greater cognitive effort in solving complex problems—we hypothesize that long-form reasoning provides greater benefits under certain conditions. This leads to a central research question: \textbf{under what circumstances is long CoT reasoning more advantageous than short CoT?} To investigate this, we perform a comprehensive three-dimensional analysis of long and short CoT strategies, evaluating their performance across different task types, levels of task difficulty, and model scales.

\paragraph{Different data types}

As illustrated in the top-left panel of Figure \ref{fig:improve}, long CoT consistently achieves higher accuracy than short CoT across a variety of datasets. The most pronounced improvements, reaching up to 20\%, occur in mathematical and knowledge-based reasoning tasks (represented by the eight datasets on the right side of the figure).



In contrast, performance gains are marginal for other datasets (the nine on the left), especially memory-based tasks like poetry and movie recommendation. Here, long CoT often introduces repetition, circular reasoning, or over-explanation, adding little value and sometimes harming performance. This suggests that long CoT’s effectiveness is task-dependent and may be unnecessary in domains with low logical reasoning demands.


\paragraph{Different difficulty levels}





Building on the analysis of task types, we next investigate the influence of task difficulty. As illustrated in Figure \ref{fig:improve}, the performance gains attributed to long CoT generally increase with task difficulty across both mathematical and knowledge reasoning datasets.

Notably, for AIME and GPQA---datasets representing the highest difficulty levels and the greatest token usage within their respective categories—improvements reach approximately 16\% and 8\%, respectively. These results underscore the significant value of long CoT in addressing complex reasoning challenges.

\paragraph{Different model scales}


To further investigate this trend, we focus on mathematics and knowledge reasoning datasets—domains where long CoT demonstrates the greatest effectiveness—and analyze performance across different model sizes (Figure \ref{fig:modelscale}). For knowledge reasoning tasks (circles), although smaller models (green) gain notable benefits from long CoT, larger models (blue and yellow) employing short CoT still outperform them by approximately 10\%. Conversely, in mathematical reasoning tasks (triangles), the performance of large models using short CoT is comparable to that of smaller models utilizing long CoT.

These results indicate that although larger models enhance knowledge reasoning by integrating information more effectively, the choice of reasoning strategy remains critical for mathematical tasks. This observation motivates the question of whether adaptively switching between long and short CoT strategies within a single large model can mitigate the trade-offs associated with scaling model size. To explore this, we propose and empirically validate a straightforward strategy selection method in the following sections.

\subsection{Impact of Token Consumption on Accuracy}


\begin{figure}[t]
    \centering
    \begin{minipage}[t]{0.4\linewidth}
        \centering
        \includegraphics[width=\linewidth]{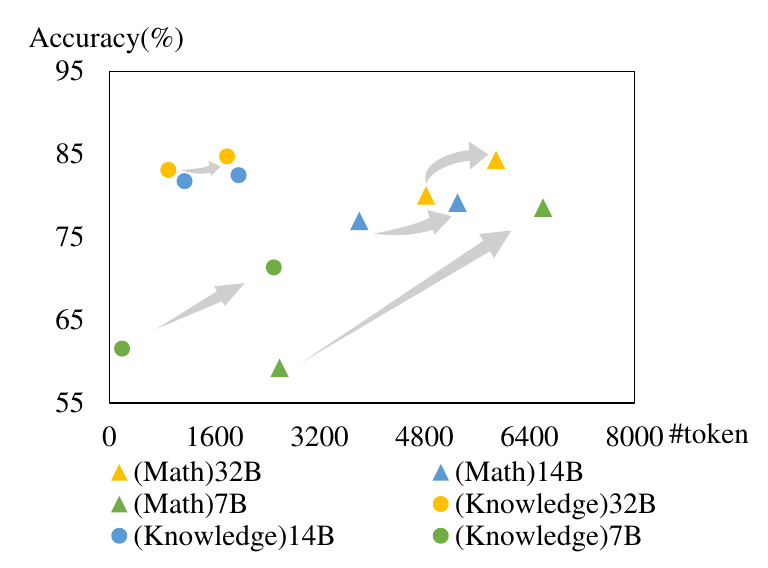}
        \caption{\textbf{Trade-off between Accuracy and Token Usage.} Each arrow indicates the change from short CoT to long CoT across datasets and DeepSeek-distilled model variants.}
        \label{fig:modelscale}
    \end{minipage}%
    \hfill
    \begin{minipage}[t]{0.55\linewidth}
        \centering
        \includegraphics[width=\linewidth]{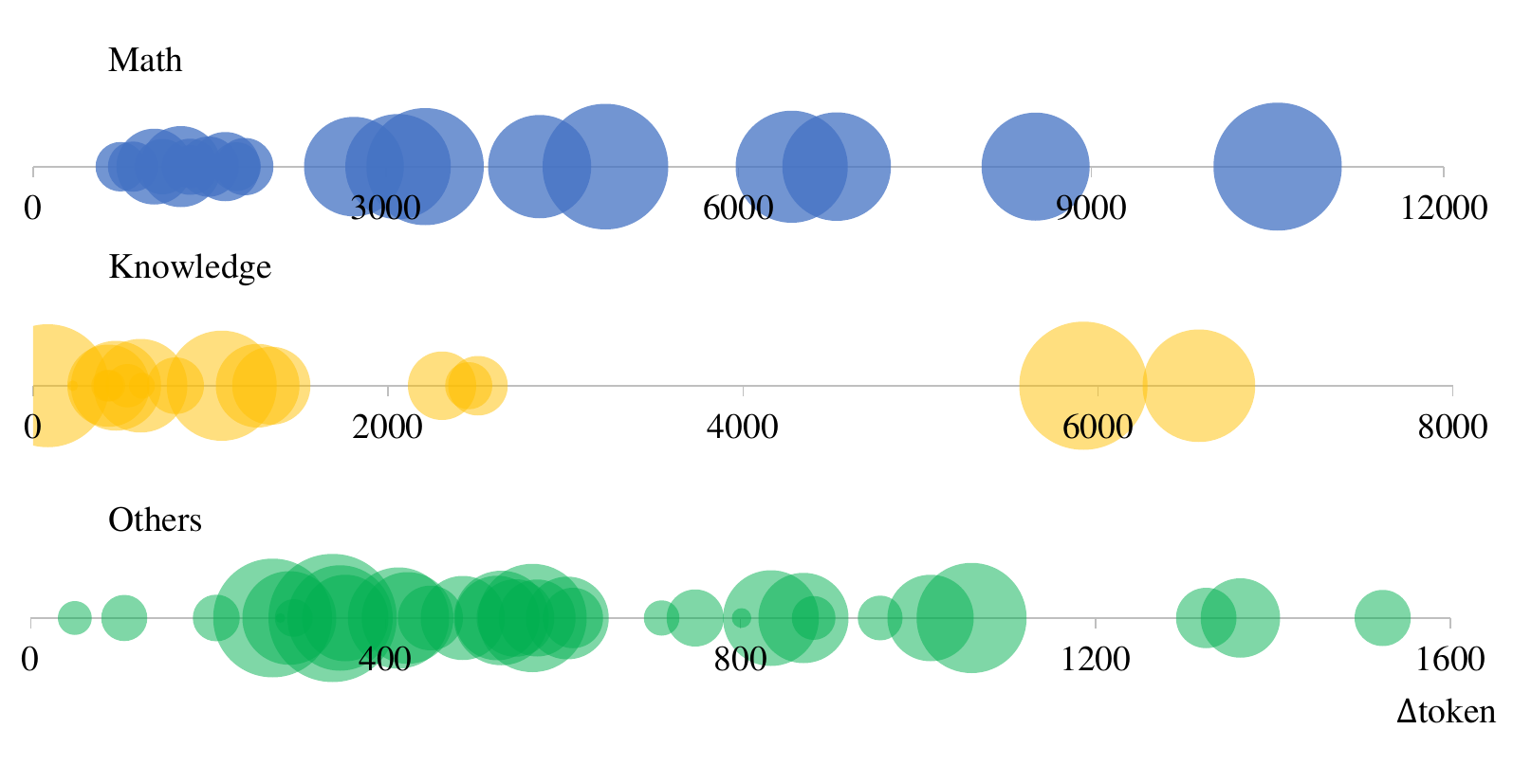}
        \caption{\textbf{Accuracy Gains vs. Token Overhead.} Each circle represents a task; larger circles indicate greater accuracy improvement. Results show that higher token usage does not always yield better performance.}
        \label{fig:bubble}
    \end{minipage}
\end{figure}



While the previous subsection highlights the effectiveness of long CoT, it also reveals considerable variation in performance gains across tasks. This prompts a crucial question: \textbf{to what extent do the accuracy improvements justify the additional token cost?} To investigate this, we analyze the cost-effectiveness of long CoT by visualizing the relationship between accuracy improvement and token overhead in a scatter plot (Figure \ref{fig:bubble}).

The plot reveals a moderate positive correlation between increased token usage and accuracy improvement in mathematical tasks, indicating that extended reasoning is particularly beneficial in this domain. In contrast, this pattern is absent in knowledge-based and other task categories, where additional token consumption does not consistently translate to performance gains.

Although most datasets experience some benefit from long CoT, the cost-effectiveness of these gains varies significantly. Furthermore, the analysis shows no clear boundary distinguishing samples with low reward but high token cost from those with high reward and comparable cost. This suggests that the trade-off between accuracy improvement and computational expense cannot be reliably predicted based solely on dataset type, task difficulty, or model scale. These findings highlight the need for a dynamic, sample-adaptive strategy to effectively decide when to employ long versus short CoT.


\section{Analysis of CoT Switch}



\subsection{Instance Selection}

The analysis in Section \ref{sec:empirical} underscores the critical need for dynamically switching between long and short CoT strategies. In this section, we present our approach to selecting the appropriate strategy, examining it from two complementary perspectives: instance-level selection and budget-aware considerations. We explore how each perspective can contribute to optimizing overall performance.

\begin{figure}
  \centering

        \includegraphics[width=0.75\textwidth]{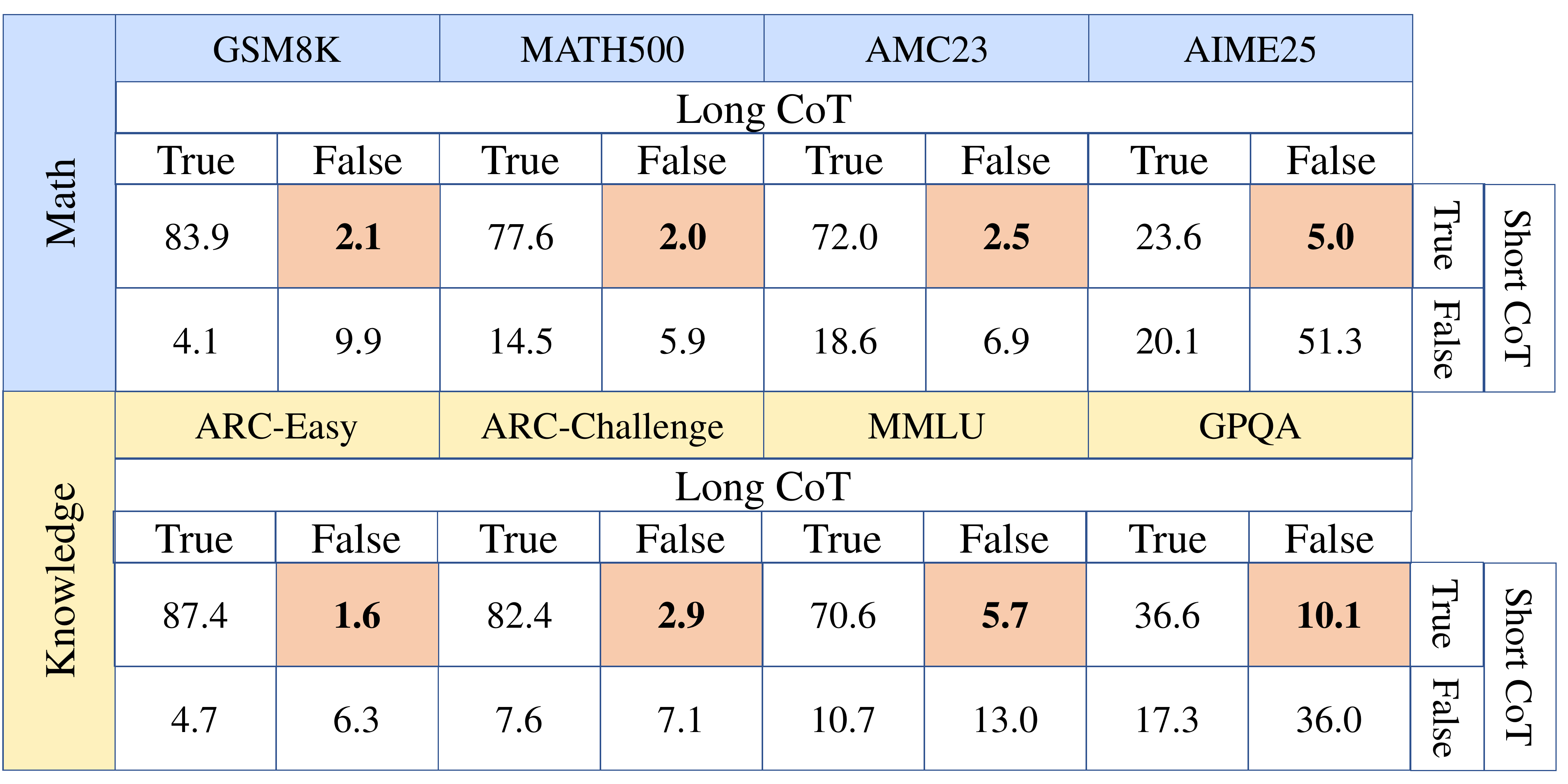}

\caption{\textbf{Confusion matrices} for various datasets, where the cells correspond to: top-left = True-True, top-right = True-False, bottom-left = False-True, and bottom-right = False-False. Here, the first value denotes the correctness of the short CoT prediction, and the second denotes that of the long CoT. The orange-highlighted top-right cells (True-False) indicate cases where long CoT harms performance, which are the primary focus of our analysis. }

  \label{fig:confusion}
\end{figure}

Recognizing that the effectiveness of long CoT varies substantially across datasets, we further investigate the factors driving this variability at the individual instance level. To this end, we introduce a confusion matrix-based analysis that quantifies the comparative performance of long and short CoT strategies on a per-instance basis. An example confusion matrix is presented in Figure \ref{fig:confusion}. Our focus lies on the True-False and False-True categories, which capture instances where the performance divergence between the two strategies is most pronounced.

As illustrated in Figure \ref{fig:confusion}, while long CoT generally improves overall performance, its benefits are not uniformly distributed across all instances—some are better addressed using short CoT. This pattern is especially pronounced in knowledge-based tasks, where long CoT can sometimes introduce irrelevant or misleading information.

Moreover, our analysis of task difficulty levels in Figure \ref{fig:confusion} shows that these performance discrepancies become increasingly significant as problem complexity rises. In other words, as tasks become more challenging, the need to adaptively switch between long and short reasoning strategies grows more critical to effectively address the specific demands of each instance.

These observations indicate that dataset type or task difficulty alone are insufficient criteria for deciding when to apply long CoT reasoning. To fully exploit the advantages of long CoT while avoiding unnecessary information in certain cases, a dynamic strategy-switching mechanism is essential—one that enables the adaptive selection of the most appropriate reasoning approach on a per-instance basis.

\subsection{Budget Aware}





\begin{figure}[t]
    \centering

    \includegraphics[width=1\linewidth]{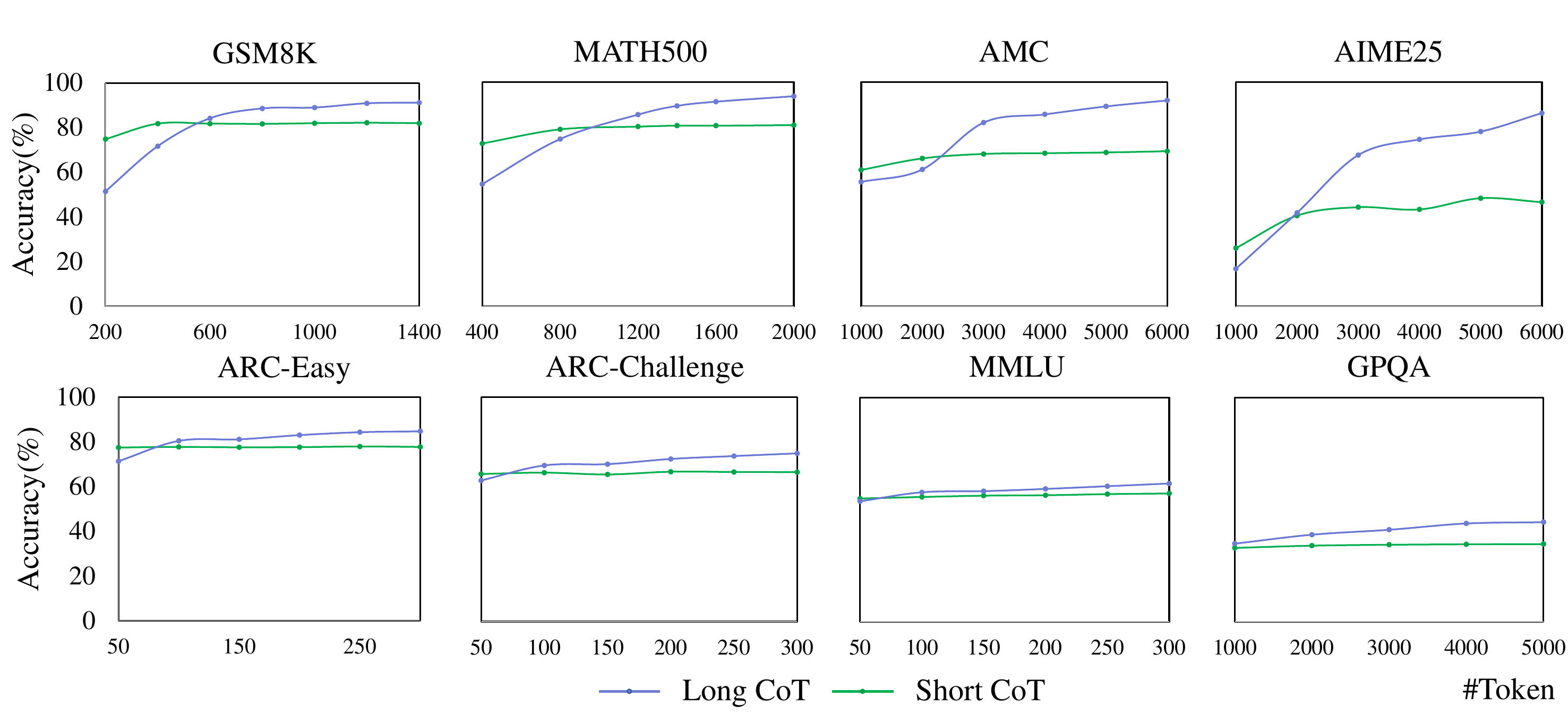}
    \caption{\textbf{Token-efficient performance of short vs. long CoT.} Short CoT excels under tight token budgets, while long CoT performs better when more tokens are available.}
    \label{fig:budget}
\end{figure}
The effectiveness of long CoT reasoning is influenced by the available computational budget, suggesting that the optimal strategy for a given instance is budget-dependent. To systematically evaluate performance under varying resource constraints, we adopt the "Budget Force" approach proposed by \citealp{Muennighoff2025s1ST}, which enforces consistent generation length limits. This enables a fair and controlled comparison between long and short CoT strategies across different token budgets.

In practice, generation is truncated once the output reaches the predefined token limit. If the reasoning process is incomplete at truncation, a closing \textit{</think>} tag is inserted to explicitly terminate the thought phase, followed by a prompt such as "Final answer" to elicit the model’s concluding response. By standardizing output length in this manner, the method ensures equitable evaluation across samples and reduces performance variability caused by differing output lengths. Experimental results under various budget settings are presented in Figure \ref{fig:budget}.

Interestingly, under tight token budgets, short CoT often achieves higher accuracy than long CoT, particularly on mathematical and simpler datasets. For example, on GSM8K with a 200-token budget, short CoT outperforms long CoT by approximately 20\%. This suggests that short CoT can yield correct answers more efficiently, avoiding the computational overhead and potential redundancy of extended reasoning. However, as the token budget increases, long CoT performance steadily improves and eventually surpasses short CoT, resulting in significant accuracy gains.

These findings underscore the promise of budget-aware strategy selection, enabling dynamic switching between long and short CoT to optimize performance within computational resource constraints.

\section{SwitchCoT: Automatic Switch between Long and Short CoT }

\begin{figure}
    \centering
    \includegraphics[width=0.9\linewidth]{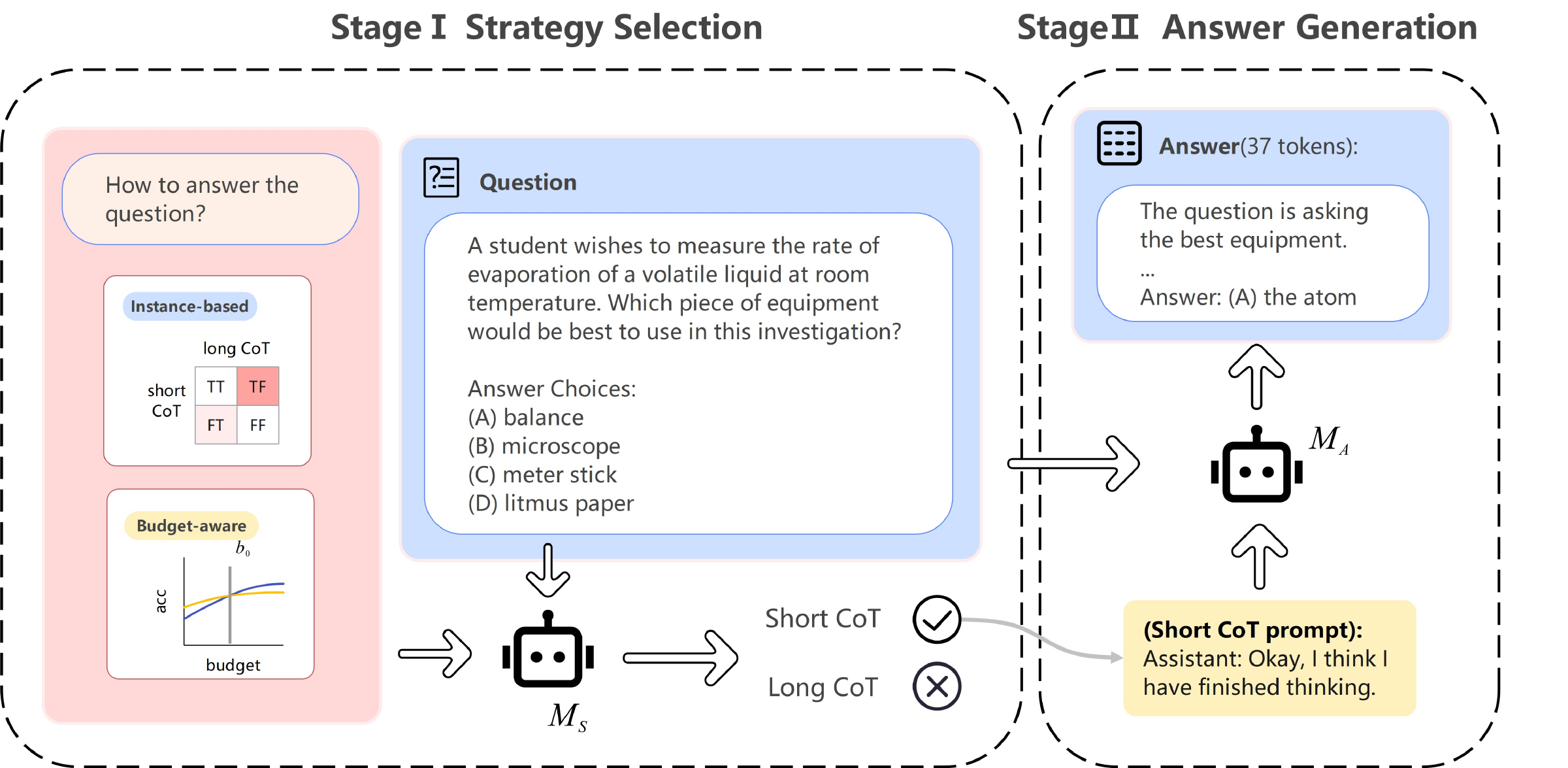}
    \caption{\textbf{The framework of SwitchCoT.} Given a question, it first selects the optimal reasoning strategy using a fine-tuned selector model, taking into account instance characteristics and, optionally, a budget constraint (Stage\ \uppercase\expandafter{\romannumeral 1}). Then in Stage\ \uppercase\expandafter{\romannumeral 2},  a generation model produces the final answer based on the selected strategy.}
    \label{fig:method}
\end{figure}


Building on the above analysis, we find that long and short Chain-of-Thought (CoT) strategies offer complementary strengths depending on the characteristics of individual instances and the available computational budget. Relying on a fixed reasoning strategy often fails to strike an effective balance between accuracy and efficiency.

To overcome this limitation, we propose SwitchCoT—a dynamic framework that adaptively selects between long and short CoT reasoning at the instance level, based on both input features and current budget constraints. This adaptive mechanism maintains high reasoning quality while avoiding unnecessary token usage. The overall architecture of our framework is illustrated in Figure~\ref{fig:method}.
\subsection{Method}

The framework of SwitchCoT consists of two stages: strategy selection and answer generation.




\paragraph{Strategy Selection}

In the first stage, the model selects the most appropriate reasoning strategy \( C \in \{C_{\texttt{short}}, C_{\texttt{long}}\} \) for a given input question \( q \).

When there are no computational budget constraints, the strategy selector \( \mathcal{M}_{S} \) predicts the optimal reasoning path solely based on the input:
\[
C = \mathcal{M}_{S}(q)
\]
This reflects the strategy expected to yield the highest performance regardless of inference cost.

To support scenarios where computational budgets are limited, we extend the selector to incorporate a budget parameter \( b \), allowing it to adaptively choose a strategy based on both the input and the available resources:
\[
C = \mathcal{M}_{S}(q, b)
\]
This unified selection mechanism is lightweight and introduces minimal computational overhead, enabling flexible deployment across a range of resource-constrained environments.

\paragraph{Answer Generation}

In the second stage, the selected strategy \( C \) is instantiated as a specific prompt template to guide the answer generation model \( \mathcal{M}_{A} \).

Specifically, the short CoT strategy \( C_{\texttt{short}} \) appends a concise marker—\textit{<think> Okay, I think I have finished thinking. </think>}—to signal minimal reasoning. In contrast, the long CoT strategy \( C_{\texttt{long}} \) prepends the token \textit{<think>} to encourage more detailed, step-by-step reasoning.

Given a question \( q \) and the selected strategy \( C \), the model generates the final output:
\[
Y = \mathcal{M}_{A}(q, C)
\]
where \( Y \) is the response conditioned on both the input and the chosen reasoning strategy.

If a computational budget \( b \) is specified, the output is truncated to fit within the token limit:
\[
Y_t = \text{Truncate}(Y, b)
\]
where $\text{Truncate}(\cdot)$ is the truncation function and $Y_t$ is a truncated version of $Y$. This ensures that the response complies with resource constraints while retaining as much meaningful content as possible.

Importantly, this stage does not require any modification to the model parameters. Instead, reasoning behavior is controlled entirely via prompting, making the framework modular, efficient, and adaptable to different tasks and budget levels.


\begin{table}[t]

  \centering
  \caption{\textbf{Main Experiment Results}. Difficulty-based strategy means that short CoT is used for GSM8K, MATH500, and ARC; long CoT is used for AMC23, AIME25, MMLU, and GPQA. TLMRE is a baseline method. SwitchCoT achieves the best performance, while significantly reducing token consumption. }
      \label{tab:exp1}
  \begin{tabular}{lcccccccc}
    \toprule
    
    \multirow{2}{*}{\textbf{Strategy}} & \multicolumn{2}{c}{\textbf{ALL}} & \multicolumn{2}{c}{\textbf{Math}} & \multicolumn{2}{c}{\textbf{Knowledge}}& \multicolumn{2}{c}{\textbf{Social}} \\
    \cmidrule(lr){2-3} \cmidrule(lr){4-5} \cmidrule(lr){6-7} \cmidrule(lr){8-9}
    & Acc (\%) & Token & Acc (\%) & Token & Acc (\%) & Token & Acc (\%) & Token \\
    \midrule
    Short CoT      & 56.2  & 73   & 73.7  & 430   & 54.9 & 46 & 48.9 & 6 \\
    \addlinespace
    Long CoT    & 88.2  & 1174   & \textbf{94.2}  & 2277  & 87.6  & 1093  & 82.3 & 559\\
     \hline
    \addlinespace
    Random   & 72.4 & 459 &  85.0 & 1112 & 71.3 & 447 & 63.5 & 238 \\
    \addlinespace
     Difficulty-based   & 78.9 & 863 &  76.3 & 553 & 82.1 & 972 & - & - \\
     \addlinespace
     TLMRE & 61.5 & 754 & 93.4 & 1448 & 56.2 & 701 & 66.6 & 486 \\
     \addlinespace
    \textbf{SwitchCoT}        & \textbf{88.9}  & 556  &  92.5  & 1333  & \textbf{88.6}  & 498  & \textbf{83.3} & 299 \\
    \bottomrule
  \end{tabular}
  \vspace{1em}
  
\end{table}

\begin{table}[t]
  \centering
  \caption{\textbf{Generalization performance results.}  SwitchCoT consistently generalizes well.  }
      \label{tab:exp2}
\begin{tabular}{lcccccc}
    \toprule
    \multirow{2}{*}{\textbf{Strategy}} & \multicolumn{2}{c}{\textbf{Fact}} & \multicolumn{2}{c}{\textbf{Creative}} & \multicolumn{2}{c}{\textbf{Sentiment}} \\
    \cmidrule(lr){2-3} \cmidrule(lr){4-5} \cmidrule(lr){6-7} 
    & Acc (\%) & Token & Acc (\%) & Token & Acc (\%) & Token \\
    \midrule
    Short CoT      & 57.4  & 7   & 59.7  & 22   & 70.1 & 30 \\

    \addlinespace
    Long CoT       & \textbf{62.1}  & 1271 & 60.2  & 1205  & 74.8  & 409\\
        \hline
    \addlinespace
    Random         & 58.6  & 322  & 59.2  & 287  & 72.7  & 196.5 \\
    \addlinespace
    \textbf{SwitchCoT}     &{60.2}  & 314  &\textbf{60.3} & 354  & \textbf{74.9}  & 154 \\
    \bottomrule
\end{tabular}
\end{table}

\section{Analysis of SwitchCoT}

Building on our proposed two-stage framework, we finetune the base model DeepSeek-R1-Distill-Qwen-7B on datasets from the mathematical, knowledge, and social domains to serve as the strategy selection model, while using the original base model for answer generation. Experimental results demonstrate that our method significantly reduces token consumption while maintaining competitive accuracy. Moreover, under constrained token budgets, \textbf{SwitchCoT} exhibits strong adaptability and delivers stable performance across a range of tasks. Further training details are provided in Appendix~\ref{app:train detail}.

\subsection{Instance-Level Selection Results}

To evaluate the effectiveness of our strategy selector at the individual instance level, we analyze its impact on both token consumption and accuracy. As baselines, we consider (i) random strategy selection, (ii) difficulty-based selection, and (iii) the method proposed by~\cite{arora2025traininglanguagemodelsreason}. 

As shown in Table~\ref{tab:exp1}, compared to using long CoT exclusively, our method reduces average token consumption from approximately 1000 to 500 tokens per instance—achieving nearly a 50\% reduction—while maintaining comparable accuracy. Furthermore, on both knowledge-intensive and social reasoning tasks, our method significantly outperforms all baseline approaches in terms of overall performance.


We further assess generalization to out-of-distribution datasets, as shown in Table~\ref{tab:exp2}. On these datasets, our method achieves accuracy comparable to long CoT while reducing token usage by more than half. For creative tasks, although our approach uses more tokens than a random baseline, it avoids the performance degradation associated with random strategy selection. These findings highlight that our method achieves a strong balance between performance and computational efficiency, even on unseen data.

\subsection{Budget-Aware Evaluation}

\begin{figure}[t]
    \centering
    \includegraphics[width=1\linewidth]{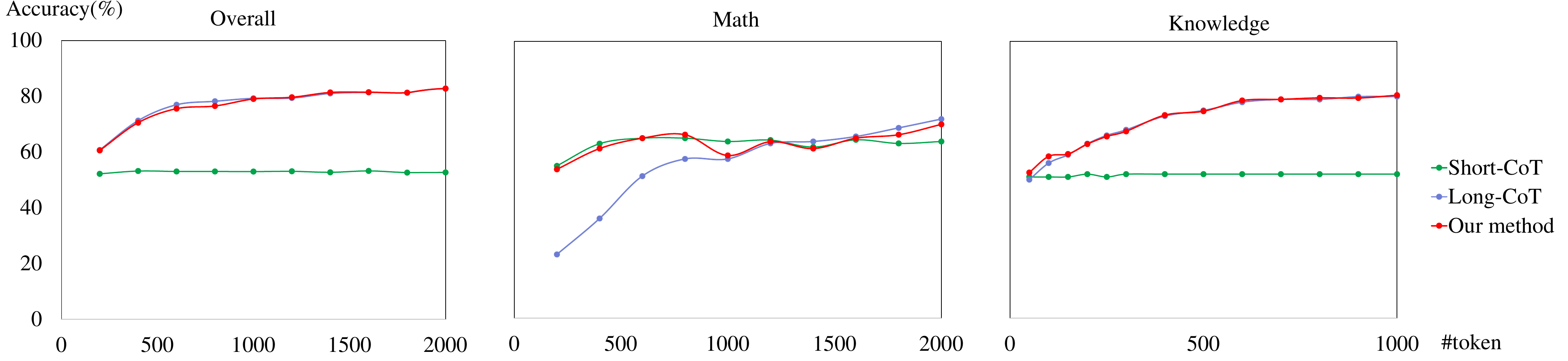}
    \caption{\textbf{Experiment results under different token budget.} The red line represents our method. It consistently achieves the best performance across different token budgets.}
    \label{fig:exp}
\end{figure}

Beyond instance-level adaptation, we evaluate our method under varying computational budget constraints. As illustrated in Figure~\ref{fig:exp}, \textbf{SwitchCoT} consistently achieves optimal or near-optimal performance across different budget levels. This demonstrates its ability to effectively combine the advantages of both short and long CoT strategies, resulting in a smooth and balanced trade-off curve that adjusts to resource availability.

These results confirm that our strategy selection model can dynamically adjust the reasoning strategy based on instance characteristics and budget constraints, achieving a practical and effective balance between accuracy and computational cost.

\section{Conclusion \& Limitation}
\label{sec:conclusion}

In this work, we present a comprehensive empirical comparison between long and short CoT reasoning strategies. Our analysis reveals that while long CoT generally improves accuracy, particularly for complex reasoning tasks, it also incurs substantial computational overhead and may yield only marginal gains in simpler scenarios. To address this trade-off, we propose SwitchCoT, a budget-aware framework that dynamically selects between short and long CoT strategies based on instance-specific needs. SwitchCoT achieves a favorable balance between reasoning effectiveness and computational efficiency, and our experimental results demonstrate its superiority over many baselines across diverse datasets and task settings. We hope this work encourages future research on adaptive reasoning mechanisms and contributes toward building more resource-efficient language models.

Despite its effectiveness, our approach has limitations. A central constraint lies in its binary formulation of reasoning strategies, where each instance is forced to choose between short or long CoT. This dichotomous setup, while operationally simple and effective for evaluation, does not fully capture the continuous nature of reasoning demands observed in real-world tasks. In practice, the required depth of reasoning often lies along a spectrum, and many instances would benefit from intermediate levels of deliberation that fall between the two extremes.
Our framework currently lacks the ability to modulate the reasoning process dynamically based on intermediate signals during generation.

Addressing these challenges calls for future work on more flexible and fine-grained reasoning control, ideally allowing models to dynamically determine not just which strategy to apply, but how much reasoning to perform. Such advancements are essential to better align reasoning depth with task complexity, ultimately leading to more capable and efficient language models.








\clearpage

\bibliography{ref}
\bibliographystyle{ref}









\appendix
\newpage

\section{Datasets}
\label{app:dataset}

Our evaluation covers a wide range of task categories, including math, factual and knowledge, social reasoning, sentiment, and memory tasks. Table~\ref{tab:task_descriptions} provides a detailed description of each task category. Table~\ref{tab:dataset_assignment} summarizes the datasets assigned to each category.


\begin{table}[ht]
  \caption{Description of our experiments dataset.}
  \label{tab:datasetsdes}
  \centering
  \begin{tabular}{cp{10cm}}
    \toprule
   \textbf{Dataset} & \multicolumn{1}{c}{\textbf{Type}} \\  
    \midrule
        Math & Tasks that require numerical reasoning, symbolic manipulation, or multi-step arithmetic calculations. These assess the model’s ability to follow logical chains of mathematical thought.  \\\addlinespace
        \hline
        \addlinespace
        Knowledge & Tasks related to academic or scientific knowledge, often involving subject-specific understanding in areas such as physics, biology, or history. These assess the model's grasp of structured disciplinary content. \\
        \addlinespace
        \hline
        \addlinespace
        Fact & Tasks focused on recalling specific factual information, such as encyclopedic knowledge. Compared to knowledge tasks, these place greater emphasis on memory precision and exactness. \\
         \addlinespace
        \hline
        \addlinespace
        Soicial & Tasks involving understanding of social dynamics, intentions, and human behavior. These require the model to reason about interpersonal contexts and social cues.   \\
         \addlinespace
        \hline
        \addlinespace
         SST & Tasks derived from the Stanford Sentiment Treebank, measuring the model’s ability to detect sentiment polarity (e.g., positive, neutral, negative) in natural language.   \\
          \addlinespace
        \hline
        \addlinespace
        Creative & Tasks that test the model’s ability to understand narrative structure, stylistic features, and contextual coherence. These include movie plot matching and poetry selection, requiring sensitivity to theme, tone, and textual alignment. \\
        
    \bottomrule
    \label{tab:task_descriptions}
  \end{tabular}
\end{table}


\begin{table}[ht]
  \caption{A full list of our experiments dataset.}
  \label{tab:datasets}
  \centering
  \begin{tabular}{cccc}
    \toprule
    Dataset &  Type &  Answer Format &  \#Questions  \\
    \midrule
    GSM8K \citep{cobbe2021training}     & math  & Short Answer   & 1319    \\
    Math-500 \citep{lightman2023lets}   & math  & Short Answer   & 500    \\
    AMC  \citep{amc23}      & math  & Short Answer   & 40   \\
    AIME25  \citep{aime24}   & math  & Short Answer   & 30   \\
    ARC-Easy \citep{allenai:arc}   & knowledge  & Multiple Choice    & 2376     \\
    ARC-Challenge \citep{allenai:arc}   & knowledge  & Multiple Choice     & 1172     \\
    MMLU   \citep{hendryckstest2021}    & knowledge  & Multiple Choice    & 9000    \\
    GPQA  \citep{rein2024gpqa}     & knowledge  & Multiple Choice     & 198     \\
    Open-hobby \citep{open} & Fact  & Multiple Choice     & 1242     \\
    Open-movie \citep{open} & Fact  & Multiple Choice     & 4313     \\
    Open-rated \citep{open} & Fact  & Multiple Choice     & 2185     \\
    Open-for-kids \citep{open}  & Fact  & Multiple Choice     & 754    \\
    SIQA  \cite{sap2019socialiqa}     & Social  & Multiple Choice     & 1954     \\   
    SST2 \citep{socher-etal-2013-recursive}& SST  & Number     & 1821     \\
    SST5 \citep{socher-etal-2013-recursive}& SST  & Number     & 2210     \\
    Poetry \citep{poetry}& Creative  & Multiple Choice     & 552     \\
    Movie recommendation \cite{srivastava2023beyond} & Creative  & Multiple Choice     & 250    \\
    \bottomrule
  \end{tabular}
  \label{tab:dataset_assignment}
\end{table}

\begin{table}[ht]
  \caption{A full list of our experiments dataset in MATH.}
  \label{tab:datasetsmath}
  \centering
  \begin{tabular}{cccccc}
    \toprule
    \multirow{2}{*}{\textbf{Dataset}} & \multirow{2}{*}{\textbf{Model}} & \multicolumn{2}{c}{\textbf{Short CoT}} & \multicolumn{2}{c}{\textbf{Long CoT}} \\
\cmidrule(lr){3-4} \cmidrule(lr){5-6}
& & Accuracy & Token & Accuracy & Token \\
    \midrule
    GSM8K   &       DeepSeek-R1-Distill-Qwen-32B & 90.7 & 438 & 94.9 & 1181 \\
GSM8K   &       DeepSeek-R1-Distill-Qwen-14B & 90.0 & 346 & 95.3 & 1445 \\
GSM8K   &       DeepSeek-R1-Distill-Qwen-7B & 80.6 & 183 & 91.8 & 1439 \\
GSM8K   &       DeepSeek-R1-Distill-Llama-8B & 81.7 & 578 & 91.5 & 1608 \\
GSM8K   &       Qwen3-8B & 87.3 & 163 & 95.5 & 1797 \\
\hline
MATH-500        &       DeepSeek-R1-Distill-Qwen-32B & 90.1 & 2181 & 93.3 & 3522 \\
MATH-500        &       DeepSeek-R1-Distill-Qwen-14B & 86.6 & 1808 & 92.2 & 3608 \\
MATH-500        &       DeepSeek-R1-Distill-Qwen-7B & 73.4 & 782 & 92.4 & 3886 \\
MATH-500        &       DeepSeek-R1-Distill-Llama-8B & 73.3 & 1616 & 90.3 & 4345 \\
MATH-500        &       Qwen3-8B & 74.5 & 703 & 92.7 & 5011 \\
\hline
AMC23   &       DeepSeek-R1-Distill-Qwen-32B & 89.2 & 4874 & 93.5 & 5723 \\
AMC23   &       DeepSeek-R1-Distill-Qwen-14B & 88.3 & 3631 & 92.4 & 5360 \\
AMC23   &       DeepSeek-R1-Distill-Qwen-7B & 62.4 & 1619 & 89.4 & 6487 \\
AMC23   &       DeepSeek-R1-Distill-Llama-8B & 65.8 & 3217 & 89.3 & 6552 \\
AMC23   &       Qwen3-8B & 66.8 & 1380 & 88.3 & 7833 \\
\hline
AIME25  &       DeepSeek-R1-Distill-Qwen-32B & 50.1 & 11806 & 55.5 & 13142 \\
AIME25  &       DeepSeek-R1-Distill-Qwen-14B & 36.7 & 9451 & 42.9 & 10945 \\
AIME25  &       DeepSeek-R1-Distill-Qwen-7B & 20.5 & 7782 & 40.6 & 14617 \\
AIME25  &       DeepSeek-R1-Distill-Llama-8B & 12.8 & 5735 & 32.8 & 14262 \\
AIME25  &       Qwen3-8B & 18.8 & 2495 & 46.9 & 13082 \\
    \bottomrule
  \end{tabular}
\end{table}

\begin{table}[ht]
  \caption{A full list of our experiments dataset in knowledge.}
  \label{tab:datasetsknow}
  \centering
  \begin{tabular}{cccccc}
    \toprule
    \multirow{2}{*}{\textbf{Dataset}} & \multirow{2}{*}{\textbf{Model}} & \multicolumn{2}{c}{\textbf{Short CoT}} & \multicolumn{2}{c}{\textbf{Long CoT}} \\
\cmidrule(lr){3-4} \cmidrule(lr){5-6}
& & Accuracy & Token & Accuracy & Token \\
    \midrule
    ARC-Easy        &       DeepSeek-R1-Distill-Qwen-32B & 94.2 & 229 & 93.9 & 549 \\
ARC-Easy        &       DeepSeek-R1-Distill-Qwen-14B & 94.0 & 330 & 93.9 & 515 \\
ARC-Easy        &       DeepSeek-R1-Distill-Qwen-7B & 80.5 & 10 & 88.4 & 616 \\
ARC-Easy        &       DeepSeek-R1-Distill-Llama-8B & 82.7 & 221 & 90.0 & 688 \\
ARC-Easy        &       Qwen3-8B & 93.7 & 61 & 94.3 & 674 \\
\hline
ARC-Challenge   &       DeepSeek-R1-Distill-Qwen-32B & 93.2 & 267 & 94.1 & 685 \\
ARC-Challenge   &       DeepSeek-R1-Distill-Qwen-14B & 93.3 & 364 & 93.4 & 588 \\
ARC-Challenge   &       DeepSeek-R1-Distill-Qwen-7B & 69.4 & 14 & 83.0 & 830 \\
ARC-Challenge   &       DeepSeek-R1-Distill-Llama-8B & 79.8 & 355 & 85.9 & 780 \\
ARC-Challenge   &       Qwen3-8B & 90.9 & 84 & 93.8 & 886 \\
\hline
MMLU    &       DeepSeek-R1-Distill-Qwen-32B & 87.7 & 465 & 89.4 & 997 \\
MMLU    &       DeepSeek-R1-Distill-Qwen-14B & 85.9 & 540 & 86.8 & 971 \\
MMLU    &       DeepSeek-R1-Distill-Qwen-7B & 59.9 & 92 & 66.2 & 1357 \\
MMLU    &       DeepSeek-R1-Distill-Llama-8B & 64.8 & 309 & 75.7 & 1373 \\
MMLU    &       Qwen3-8B & 83.0 & 117 & 88.5 & 1459 \\
\hline
GPQA    &       DeepSeek-R1-Distill-Qwen-32B & 57.3 & 2627 & 61.5 & 4932 \\
GPQA    &       DeepSeek-R1-Distill-Qwen-14B & 53.8 & 3342 & 55.8 & 5797 \\
GPQA    &       DeepSeek-R1-Distill-Qwen-7B & 36.3 & 643 & 47.7 & 7213 \\
GPQA    &       DeepSeek-R1-Distill-Llama-8B & 42.5 & 4275 & 45.7 & 6782 \\
GPQA    &       Qwen3-8B & 43.8 & 517 & 58.6 & 6436 \\
    \bottomrule
  \end{tabular}
\end{table}

\begin{table}[ht]
  \caption{A full list of our experiments dataset in others.}
  \label{tab:datasetsothers}
  \centering
  \begin{tabular}{cccccc}
    \toprule
    \multirow{2}{*}{\textbf{Dataset}} & \multirow{2}{*}{\textbf{Model}} & \multicolumn{2}{c}{\textbf{Short CoT}} & \multicolumn{2}{c}{\textbf{Long CoT}} \\
\cmidrule(lr){3-4} \cmidrule(lr){5-6}
& & Accuracy & Token & Accuracy & Token \\
    \midrule
    Open-hobby      &       DeepSeek-R1-Distill-Qwen-32B & 78.2 & 187 & 82.0 & 674 \\
Open-hobby      &       DeepSeek-R1-Distill-Qwen-14B & 79.8 & 384 & 79.5 & 679 \\
Open-hobby      &       DeepSeek-R1-Distill-Qwen-7B & 46.0 & 9 & 52.3 & 1070 \\
Open-hobby      &       DeepSeek-R1-Distill-Llama-8B & 62.9 & 372 & 66.6 & 898 \\
Open-hobby      &       Qwen3-8B & 73.2 & 117 & 77.4 & 988 \\
\hline
Open-movie      &       DeepSeek-R1-Distill-Qwen-32B & 61.0 & 185 & 64.1 & 731 \\
Open-movie      &       DeepSeek-R1-Distill-Qwen-14B & 61.6 & 377 & 62.3 & 673 \\
Open-movie      &       DeepSeek-R1-Distill-Qwen-7B & 34.4 & 8 & 36.0 & 1532 \\
Open-movie      &       DeepSeek-R1-Distill-Llama-8B & 52.8 & 276 & 56.4 & 880 \\
Open-movie      &       Qwen3-8B & 59.0 & 108 & 60.0 & 1066 \\
\hline
Open-rated      &       DeepSeek-R1-Distill-Qwen-32B & 73.0 & 175 & 75.2 & 625 \\
Open-rated      &       DeepSeek-R1-Distill-Qwen-14B & 73.3 & 333 & 73.4 & 615 \\
Open-rated      &       DeepSeek-R1-Distill-Qwen-7B & 42.8 & 9 & 46.0 & 1373 \\
Open-rated      &       DeepSeek-R1-Distill-Llama-8B & 63.0 & 274 & 66.0 & 845 \\
Open-rated      &       Qwen3-8B & 66.8 & 89 & 71.7 & 923 \\
\hline
Open-for-kids   &       DeepSeek-R1-Distill-Qwen-32B & 70.9 & 159 & 75.6 & 689 \\
Open-for-kids   &       DeepSeek-R1-Distill-Qwen-14B & 71.9 & 364 & 73.0 & 573 \\
Open-for-kids   &       DeepSeek-R1-Distill-Qwen-7B & 43.9 & 8 & 45.8 & 1333 \\
Open-for-kids   &       DeepSeek-R1-Distill-Llama-8B & 60.1 & 325 & 60.8 & 1036 \\
Open-for-kids   &       Qwen3-8B & 63.8 & 88 & 67.7 & 1102 \\
\hline
SIQA    &       DeepSeek-R1-Distill-Qwen-32B & 79.6 & 39 & 78.8 & 488 \\
SIQA    &       DeepSeek-R1-Distill-Qwen-14B & 77.2 & 310 & 77.7 & 482 \\
SIQA    &       DeepSeek-R1-Distill-Qwen-7B & 59.9 & 8 & 65.3 & 595 \\
SIQA    &       DeepSeek-R1-Distill-Llama-8B & 64.6 & 128 & 69.9 & 553 \\
SIQA    &       Qwen3-8B & 75.0 & 84 & 77.7 & 884 \\
\hline
SST2    &       DeepSeek-R1-Distill-Qwen-32B & 90.6 & 103 & 94.1 & 301 \\
SST2    &       DeepSeek-R1-Distill-Qwen-14B & 91.8 & 255 & 92.4 & 305 \\
SST2    &       DeepSeek-R1-Distill-Qwen-7B & 83.4 & 32 & 87.3 & 387 \\
SST2    &       DeepSeek-R1-Distill-Llama-8B & 77.7 & 86 & 85.1 & 359 \\
SST2    &       Qwen3-8B & 88.5 & 48 & 93.2 & 339 \\
\hline
SST5    &       DeepSeek-R1-Distill-Qwen-32B & 41.0 & 154 & 47.2 & 611 \\
SST5    &       DeepSeek-R1-Distill-Qwen-14B & 45.1 & 334 & 46.2 & 439 \\
SST5    &       DeepSeek-R1-Distill-Qwen-7B & 34.6 & 39 & 40.0 & 454 \\
SST5    &       DeepSeek-R1-Distill-Llama-8B & 26.6 & 135 & 35.2 & 475 \\
SST5    &       Qwen3-8B & 43.0 & 92 & 44.9 & 703 \\
\hline
Poetry  &       DeepSeek-R1-Distill-Qwen-32B & 81.5 & 335 & 81.4 & 683 \\
Poetry  &       DeepSeek-R1-Distill-Qwen-14B & 79.8 & 398 & 78.5 & 729 \\
Poetry  &       DeepSeek-R1-Distill-Qwen-7B & 49.2 & 150 & 51.0 & 899 \\
Poetry  &       DeepSeek-R1-Distill-Llama-8B & 63.1 & 144 & 69.3 & 709 \\
Poetry  &       Qwen3-8B & 79.7 & 146 & 79.0 & 959 \\
\hline
MovieRec        &       DeepSeek-R1-Distill-Qwen-32B & 57.8 & 131 & 58.0 & 932 \\
MovieRec        &       DeepSeek-R1-Distill-Qwen-14B & 57.4 & 490 & 61.8 & 915 \\
MovieRec        &       DeepSeek-R1-Distill-Qwen-7B & 47.0 & 8 & 41.0 & 1959 \\
MovieRec        &       DeepSeek-R1-Distill-Llama-8B & 47.4 & 599 & 53.2 & 948 \\
MovieRec        &       Qwen3-8B & 66.8 & 88 & 67.8 & 971 \\
    \bottomrule
  \end{tabular}
\end{table}

\section{Empirical Analysis Result Details}

Our empirical results in section\ref{sec:empirical} can be found in Table \ref{tab:datasetsmath}, Table \ref{tab:datasetsknow} and Table \ref{tab:datasetsothers}.




\section{Training Details}
\label{app:train detail}

We begin with a pre-trained reasoning language model $\mathcal{M}_0$ and a collection of reasoning datasets $\mathcal{D} = \{ \mathcal{D}_1, \mathcal{D}_2, \dots, \mathcal{D}_M \}$, where each dataset $\mathcal{D}_i = \{(x_j, y_j^{\text{gold}})\}_{j=1}^{N_i}$ contains input prompts and corresponding final answers. Our goal is to annotate each instance with a reasoning strategy(short or long CoT) using two complementary approaches: instance-based and budget-aware selection.

\paragraph{Instance-Based Labeling} 
Given a question $q\in \mathcal{D}$, we define two prompting functions: $\pi_S$ and $\pi_L$, corresponding to short and long CoT strategies, respectively. Each function generates an answer:
$
A_S = \pi_S(q)$ and $ A_L = \pi_L(q).
$
We evaluate the outputs $A_S$ and $A_L$ using the confusion matrix. Based on the evaluation, each question is assigned to one of four classes:

\[
(\texttt{True-True}) \rightarrow \texttt{short}, \quad
(\texttt{True-False}) \rightarrow \texttt{short}, \quad\]
\[
(\texttt{False-True}) \rightarrow \texttt{long}, \quad
(\texttt{False-False}) \rightarrow \texttt{discard}
\]

If it falls into the False-False region, we regard such questions as inherently difficult for the base model to handle effectively. Keeping these samples may interfere with the training of the strategy selector. Therefore, we discard all samples in this region.
\paragraph{Budget-Aware Labeling}
Following our empirical analysis in Section~\ref{sec:empirical}, we define a dataset-specific budget threshold $\tau_i$ that separates short and long CoT strategies. For each instance $q \in \mathcal{D}_i$, we sample a set of budget values from the interval $[0,\, 2b_d]$:
\[
b_j \in \left\{ b_d + \delta_1,\, b_d + \delta_2,\, \dots,\, b_d + \delta_{10} \right\},
\]
where $\delta_i$ are uniformly sampled offsets. Each sampled budget $b_x$ determines the optimal strategy based on which CoT variant satisfies the budget constraint while maintaining accuracy.

\[
\mathcal{D}_{\text{final}} = 
\underbrace{\left\{\, (x, \text{strategy}(x)) \mid x \in \mathcal{D} \,\right\}}_{\text{Instance-Based}} 
\cup 
\underbrace{\left\{\, (x, b_x, \text{strategy}(x)) \mid x \in \mathcal{D} \,\right\}}_{\text{Budget-Based}}
\]

\paragraph{Strategy Prediction Training}
After labeling each question using the sample-based and budget-aware methods, we train the model $\mathcal{M}_0$ to predict the most suitable CoT reasoning strategy for a given question.

Specifically, for each instance, either in the form $(q, s^*)$ or $(q, b, s^*)$, where $q$ is the input question, $b$ is the associated budget value (if applicable), and $s^* \in \{\texttt{short CoT}, \texttt{long CoT}\}$ is the ground-truth strategy, the model $\mathcal{M}_0$ predicts the strategy via:
\[
\hat{s} = \mathcal{M}_0(q) \quad \text{or} \quad \hat{s} = \mathcal{M}_0(q, b),
\]
depending on whether budget information is included.

The model is trained by minimizing the cross-entropy loss between the predicted distribution and the ground-truth label:
\[
\mathcal{L}_{\text{strategy}} = - \sum_{i=1}^{N} \log P_{\mathcal{M}_0}(s_i^* \mid q_i, b_i),
\]
where $P_{\mathcal{M}_0}(s \mid q, b)$ denotes the predicted probability of strategy $s$ given input $q$ and budget $b$. For instances without budget information, the model conditions only on $q$.


This training procedure yields a strategy selection model $\mathcal{M}'$, which serves as an efficient and lightweight module for reasoning strategy selection at inference time.

\clearpage

\section{Example Prompts}

We list a few prompt response pairs here containing different type of datasets.

\begin{tcolorbox}[
    breakable,
    colback=ForestGreen!10,     
    colframe=ForestGreen!80!black, 
    title=Strategy selection example by DeepSeek-R1-Distill-Qwen-7B from ARC-Challenge,       
    arc=4mm,              
    boxrule=1pt,          
    fonttitle=\bfseries,  
    coltitle=white        
]
\textbf{Prompt}: \verb+<|User|>+You are given a question. Decide the most appropriate method to answer it based on the complexity and depth required.\\
Question: Molly was making iced tea by putting tea bags in boiling water. She needed to cool the hot tea quickly, so she poured it into a pitcher full of ice. Which best describes the flow of thermal energy? Answer Choices: (A) from the ice to the tea (B) from the tea to the ice (C) from the pitcher to the tea (D) from the ice to the pitcher.\\
Which method should be used to answer this question?\\
A. Short Chain of Thought – Use this if the question is simple or factual, and can be answered correctly without deep reasoning or extended explanation.\\
B. Long Chain of Thought – Use this if the question requires in-depth reasoning, multiple steps, or a comprehensive explanation to reach a correct answer.\\
Just answer A or B. Do not answer the question.\\
\verb+<|Assistant|>+\verb+<think>+Okay, I think I have finished thinking.\verb+</think>+
\\\\
\textbf{Response}:
B. Long Chain of Thought
\end{tcolorbox}

\begin{tcolorbox}[
    breakable,
    colback=blue!5,       
    colframe=blue!75!black, 
    title=A Long-CoT DeepSeek-R1-Distill-Qwen-7B example from ARC-Challenge,       
    arc=4mm,              
    boxrule=1pt,          
    fonttitle=\bfseries,  
    coltitle=white        
]

\textbf{Prompt}: \verb+<|User|>+Molly was making iced tea by putting tea bags in boiling water. She needed to cool the hot tea quickly, so she poured it into a pitcher full of ice. Which best describes the flow of thermal energy? Answer Choices: (A) from the ice to the tea (B) from the tea to the ice (C) from the pitcher to the tea (D) from the ice to the pitcher.\\
Think step by step and then finish your answer with 'the answer is (X)' where X is the correct letter choice.\\
\verb+<|Assistant|>+\verb+<think>+
\\\\
\textbf{Ground Truth}: (B)\\\\
\textbf{Response}:
Okay, so I'm trying to figure out this question about Molly making iced tea. Let me read it again. She was putting tea bags in boiling water, and then she poured it into a pitcher full of ice. The question is about the flow of thermal energy, and the options are whether it's from the ice to the tea, tea to ice, pitcher to tea, or ice to pitcher.\\
Hmm, I remember that thermal energy flows from a hotter object to a cooler object. So first, I need to figure out which is hotter: the tea or the ice.\\
Molly is using boiling water, which I know boils at 100 degrees Celsius. So the tea is pretty hot. Then she pours it into a pitcher full of ice. Ice is at 0 degrees Celsius, which is much cooler. So the tea is definitely hotter than the ice.\\
When two objects are in contact, heat moves from the hotter to the colder. So the tea, being hotter, should transfer heat to the ice. But wait, the pitcher is involved too. The pitcher is holding the ice, so the heat from the tea would go into the ice. The ice would absorb the heat and melt, right? And as the ice melts, it becomes water, which is at 0 degrees, so it can't transfer much heat anymore because it's at the same temperature as the surrounding water.\\
Wait, but the pitcher is just holding the ice. So the tea is in the pitcher, right? So the tea is at 100 degrees, the ice at 0. So the tea would lose heat to the ice. But looking at the options, option B is from the tea to the ice. Option A is from ice to tea, which doesn't make sense because ice is colder. So the correct flow should be from tea to ice, which is option B.\\
I think I got it. The tea is hotter, so it loses heat to the ice, which melts. The pitcher is just the container, but the key is the tea and the ice in contact. So the thermal energy flows from tea to ice.\verb|</think>|\\
The flow of thermal energy is from the hotter tea to the cooler ice, as heat transfers from a higher temperature to a lower one.\\
the answer is (B)
\end{tcolorbox}

\begin{tcolorbox}[
    breakable,
    colback=BrickRed!10,
    colframe=BrickRed!80!black,
    title=A Short-CoT DeepSeek-R1-Distill-Qwen-7B example from ARC-Challenge,       
    arc=4mm,              
    boxrule=1pt,          
    fonttitle=\bfseries,  
    coltitle=white        
]
\textbf{Prompt}: \verb+<|User|>+Molly was making iced tea by putting tea bags in boiling water. She needed to cool the hot tea quickly, so she poured it into a pitcher full of ice. Which best describes the flow of thermal energy?\\ Answer Choices: (A) from the ice to the tea (B) from the tea to the ice (C) from the pitcher to the tea (D) from the ice to the pitcher.\\
Please answer the question directly without any thinking process and then finish your answer with 'the answer is (X)' where X is the correct letter choice.\\
\verb+<|Assistant|>+\verb+<think>+Okay, I think I have finished thinking.\verb+</think>+
\\\\
\textbf{Ground Truth}: (B)\\\\
\textbf{Response}:
The answer is (A).
\end{tcolorbox}


\begin{tcolorbox}[
    breakable,
    colback=ForestGreen!10,     
    colframe=ForestGreen!80!black, 
    title=Strategy selection example by DeepSeek-R1-Distill-Qwen-7B from MATH500,       
    arc=4mm,              
    boxrule=1pt,          
    fonttitle=\bfseries,  
    coltitle=white        
]
\textbf{Prompt}: \verb+<|User|>+You are given a question. Decide the most appropriate method to answer it based on the complexity and depth required.\\
Question: In how many ways can 8 people sit around a round table if 3 of the people -- Pierre, Rosa, and Thomas -- all want to sit together?  (Two seatings are considered the same if one is a rotation of the other.)\\
Which method should be used to answer this question?\\
A. Short Chain of Thought – Use this if the question is simple or factual, and can be answered correctly without deep reasoning or extended explanation.\\
B. Long Chain of Thought – Use this if the question requires in-depth reasoning, multiple steps, or a comprehensive explanation to reach a correct answer.\\
Just answer A or B. Do not answer the question.\\
\verb+<|Assistant|>+\verb+<think>+Okay, I think I have finished thinking.\verb+</think>+
\\\\
\textbf{Response}:
B. Long Chain of Thought
\end{tcolorbox}

\begin{tcolorbox}[
    breakable,
    colback=blue!5,       
    colframe=blue!75!black, 
    title=A Long-CoT DeepSeek-R1-Distill-Qwen-7B example from MATH-500,       
    arc=4mm,              
    boxrule=1pt,          
    fonttitle=\bfseries,  
    coltitle=white        
]

\textbf{Prompt}: \verb+<|User|>+In how many ways can 8 people sit around a round table if 3 of the people -- Pierre, Rosa, and Thomas -- all want to sit together?  (Two seatings are considered the same if one is a rotation of the other.)\\
Think step by step and put your final answer within \verb|\boxed{}|.\\
\verb+<|Assistant|>+\verb+<think>+
\\\\
\textbf{Ground Truth}: 720\\\\

\textbf{Response}:
Okay, so I need to figure out how many ways 8 people can sit around a round table, but with the condition that Pierre, Rosa, and Thomas all want to sit together. Hmm, circular permutations can be a bit tricky, but let me try to break it down step by step.\\
First, I remember that when arranging people around a round table, the number of arrangements is different from arranging them in a straight line because rotations are considered the same. For a circular table with n people, the number of distinct seating arrangements is (n-1)! instead of n!. So, for 8 people, it would normally be 7! ways. But that's without any restrictions.\\
But here, Pierre, Rosa, and Thomas want to sit together. So, I think I need to treat them as a single unit or "block." If I do that, instead of having 8 individual people, I have this one block plus the other 5 people. That makes a total of 6 units to arrange around the table.\\
Wait, so if I have 6 units, the number of ways to arrange them around a circular table would be (6-1)! which is 5!. But hold on, within this block, Pierre, Rosa, and Thomas can themselves be arranged in different ways. Since there are 3 people in the block, the number of ways to arrange them is 3!.\\
So, to get the total number of arrangements, I should multiply the number of ways to arrange the blocks by the number of ways to arrange the people within the block. That would be 5! multiplied by 3!.\\
Let me write that down to visualize it better:\\
Total arrangements = (Number of arrangements of blocks) × (Number of arrangements within the block)\\
Number of arrangements of blocks = (6-1)! = 5!\\
Number of arrangements within the block = 3!\\
Therefore, total arrangements = 5! × 3!\\
Calculating that, 5! is 120 and 3! is 6, so 120 × 6 is 720.\\
Wait, but hold on a second. Is there something I'm missing here? Let me think. Since the table is circular, do I need to adjust for any other symmetries? I don't think so because we already accounted for the circular nature by using (n-1)! for the blocks.\\
But let me double-check. If all 8 people were free to sit anywhere, it's 7!. If Pierre, Rosa, and Thomas are together, we treat them as a single block, so 6 units, which gives 5! arrangements. Then, within the block, 3! arrangements. So, 5! × 3! = 120 × 6 = 720. That seems right.\\
Alternatively, maybe I can think of it another way. Imagine fixing the position of the block of Pierre, Rosa, and Thomas to eliminate rotational symmetry. If I fix their position, then we can arrange the remaining 5 people. But then, within their block, they can be arranged in 3! ways. The remaining 5 people can be arranged in 5! ways. So, that would be 5! × 3! = 720 as well. So, same result.\\
Wait, another way: suppose I don't fix their position but consider all possible rotations. Since in circular permutations, fixing one point can help avoid counting rotations multiple times. So, by fixing the block, we eliminate the rotational symmetry related to their position, and then arrange the rest.\\
But I think the first method I used is correct. Treat the trio as a single block, leading to 6 units, so 5! arrangements, multiplied by 3! for the internal arrangements of the block. So, 120 × 6 = 720.\\
Just to make sure, let me consider a smaller case. Suppose there are 4 people, with 2 who want to sit together. How would that work?\\
In that case, treating the two as a block, we have 3 units: the block and the other two individuals. The number of circular arrangements would be (3-1)! = 2! = 2. Then, within the block, 2! = 2 arrangements. So, total is 2 × 2 = 4.\\
Let me list them to check:\\
Suppose people are A, B, C, D, with A and B wanting to sit together.\\
Possible arrangements:\\
1. AB, C, D: AB can be in two orders: AB or BA. So, AB C D, AB D C, BA C D, BA D C. But wait, in circular arrangements, fixing AB's position, C and D can be arranged in two ways. So, for each block position, two arrangements. Since AB can be in two orientations, that's 2 × 2 = 4. Which matches the calculation.\\
So, in the smaller case, the method works. Therefore, applying it to the original problem, 8 people with 3 wanting to sit together:\\
Number of ways = (6-1)! × 3! = 5! × 6 = 120 × 6 = 720.\\
So, I think that's the answer.\\
\textbf{Final Answer} The number of ways is \boxed{720}. \verb|</think>|\\
To determine the number of ways 8 people can sit around a round table with the condition that Pierre, Rosa, and Thomas all want to sit together, we can follow these steps:\\
1. Treat Pierre, Rosa, and Thomas as a single unit or "block." This reduces the problem to arranging 6 units (the block plus the other 5 people).\\
2. The number of ways to arrange 6 units around a circular table is (6-1)!, which is 5!.\\
3. Within the block, Pierre, Rosa, and Thomas can be arranged in 3! ways.\\
4. Multiply the number of ways to arrange the blocks by the number of ways to arrange the people within the block.\\
Calculations:
\begin{itemize}
    \item Number of ways to arrange the blocks: 5!=120
    \item Number of ways to arrange within the block: 3!=6
    \item Total number of arrangements: 5! $\times$ 3! = 120 $\times$ 6 = 720
    \item Thus, the number of ways is \boxed{720}.
\end{itemize}

\end{tcolorbox}

\begin{tcolorbox}[
    breakable,
    colback=BrickRed!10,
    colframe=BrickRed!80!black,
    title=A Short-CoT DeepSeek-R1-Distill-Qwen-7B example from MATH-500,       
    arc=4mm,              
    boxrule=1pt,          
    fonttitle=\bfseries,  
    coltitle=white        
]
\textbf{Prompt}: \verb+<|User|>+In how many ways can 8 people sit around a round table if 3 of the people -- Pierre, Rosa, and Thomas -- all want to sit together?  (Two seatings are considered the same if one is a rotation of the other.)\\
Please write the answer for this math problem directly without any thinking process.\\
\verb+<|Assistant|>+\verb+<think>+Okay, I think I have finished thinking.\verb+</think>+
\\\\
\textbf{Ground Truth}: 720\\\\
\textbf{Response}:
To determine the number of ways 8 people can sit around a round table with the condition that Pierre, Rosa, and Thomas must all sit together, we can treat the trio as a single entity or "block." This reduces the problem to arranging 6 individual people plus 1 block, totaling 7 units around the table.\\
In a circular arrangement, the number of ways to arrange $n$ units is $(n-1)!$. Therefore, the number of ways to arrange these 7 units is $(7-1)! = 6! = 720$.\\
However, within the block, Pierre, Rosa, and Thomas can be arranged among themselves in $3! = 6$ ways.\\
Multiplying these two results gives the total number of arrangements:  720 $\times$ 6 = 4320.\\
\textbf{Final Answer} The number of ways is \boxed{4320}.
\end{tcolorbox}


\begin{tcolorbox}[
    breakable,
    colback=ForestGreen!10,     
    colframe=ForestGreen!80!black, 
    title=Strategy selection example by Qwen3-8B from SST,       
    arc=4mm,              
    boxrule=1pt,          
    fonttitle=\bfseries,  
    coltitle=white        
]
\textbf{Prompt}: \verb+<|User|>+You are given a question. Decide the most appropriate method to answer it based on the complexity and depth required.\\
Question: no movement, no yuks, not much of anything.\\
Please analyze the sentiment using 0-4. 0 means very negative while 4 means very positive.\\
Which method should be used to answer this question?\\
A. Short Chain of Thought – Use this if the question is simple or factual, and can be answered correctly without deep reasoning or extended explanation.\\
B. Long Chain of Thought – Use this if the question requires in-depth reasoning, multiple steps, or a comprehensive explanation to reach a correct answer.\\
Just answer A or B. Do not answer the question.\\
\verb+<|Assistant|>+\verb+<think>+Okay, I think I have finished thinking.\verb+</think>+
\\\\
\textbf{Response}:
A. short chain of Thought
\end{tcolorbox}

\begin{tcolorbox}[
    breakable,
    colback=blue!5,       
    colframe=blue!75!black, 
    title=A Long-CoT Qwen3-8B example from SST,       
    arc=4mm,              
    boxrule=1pt,          
    fonttitle=\bfseries,  
    coltitle=white        
]

\textbf{Prompt}: \verb+<|User|>+no movement, no yuks, not much of anything.\\
Please analyze the sentiment using 0-4. 0 means very negative while 4 means very positive.\\
\verb+<|Assistant|>+\verb+<think>+
\\\\
\textbf{Ground Truth}: 1
\\\\
\textbf{Response}:
Okay, let's see. The user provided the input: "no movement, no yuks, not much of anything." and wants me to analyze the sentiment on a scale from 0 to 4, where 0 is very negative and 4 is very positive.\\
First, I need to understand the sentiment of the given statement. The phrase "no movement" could imply a lack of action or progress, which might be seen as negative. "No yuks" – I'm not sure what "yuks" refers to here. Maybe it's a typo for "jokes" or "laughs"? If it's about humor, then "no yuks" might mean there's no humor, which could be negative. "Not much of anything" suggests that there's very little happening or being offered, which is also a negative sentiment. \\
Putting it all together, the user seems to be expressing dissatisfaction or disappointment. The absence of movement, humor, and substance could indicate that the experience or product being discussed is uneventful or unimpressive. Since all the elements mentioned are negative, the sentiment is likely on the lower end of the scale. However, I need to check if there's any sarcasm or context I'm missing. But without more information, it's safer to assume a negative tone. \\On the scale from 0 to 4, where 0 is very negative and 4 is very positive, this statement would probably be a 0 or 1. But maybe 1 is too positive. Let me think again. "No movement" and "not much of anything" are definitely negative. "No yuks" could be a negative if it's about humor. So, the sentiment is very negative, so 0. But sometimes people might use sarcasm, but I don't think that's the case here. So, the answer should be 0.\verb|</think>|\\
The sentiment of the statement "no movement, no yuks, not much of anything" is \textbf{0 (very negative)}.  \\
The phrases convey dissatisfaction, boredom, or disappointment, emphasizing a lack of engagement, humor, or substance. The tone is clearly negative with no positive elements present.

\end{tcolorbox}

\begin{tcolorbox}[
    breakable,
    colback=BrickRed!10,
    colframe=BrickRed!80!black,
    title=A Short-CoT Qwen3-8B example from SST,       
    arc=4mm,              
    boxrule=1pt,          
    fonttitle=\bfseries,  
    coltitle=white        
]
\textbf{Prompt}:\verb+<|User|>+no movement, no yuks, not much of anything.\\
Please analyze the sentiment using 0-4. 0 means very negative while 4 means very positive.\\
\verb+<|Assistant|>+\verb+<think>+Okay, I think I have finished thinking.\verb|</think>|
\\\\
\textbf{Ground Truth}: 1
\\\\
\textbf{Response}:
The sentiment of the statement "no movement, no yuks, not much of anything" is likely \textbf{1} (somewhat negative). \\
The language used conveys a sense of boredom, lack of excitement, or disappointment. Words like "no movement" and "not much of anything" suggest a lack of engagement or interest, which leans toward a negative tone. However, it's not extremely negative, so a score of 1 is appropriate.
\end{tcolorbox}

\end{document}